\def\year{2022}\relax
\documentclass[letterpaper]{article} 
\usepackage{aaai22}  
\usepackage{times}  
\usepackage{helvet}  
\usepackage{courier}  
\usepackage[hyphens]{url}  
\usepackage{graphicx} 
\usepackage{natbib}  
\usepackage{caption} 
\DeclareCaptionStyle{ruled}{labelfont=normalfont,labelsep=colon,strut=off} 
\usepackage{algorithm}
\usepackage{algorithmic}

\usepackage{xcolor}

\usepackage{newfloat}
\usepackage{listings}
\floatstyle{ruled}
\newfloat{listing}{tb}{lst}{}
\floatname{listing}{Listing}
\title{Taxonomy-based \textsc{CheckList} for Large Language Model Evaluation}
\author{
    Damin Zhang}
\affiliations{
    Purdue University\\
    Computer and Information Technology\\
    zhan4060@purdue.edu
}

\begin{document}

\maketitle

\begin{abstract}
As large language models (LLMs) have been used in many downstream tasks, the internal stereotypical representation may affect the fairness of the outputs. In this work, we introduce human knowledge into natural language interventions and study pre-trained language models' (LMs) behaviors within the context of gender bias. Inspired by \textsc{CheckList} behavioral testing \cite{ribeiro2020beyond}, we present a checklist-style task that aims to probe and quantify LMs' unethical behaviors through question-answering (QA). We design three comparison studies to evaluate LMs from four aspects: \textit{consistency}, \textit{biased tendency}, \textit{model preference}, and \textit{gender preference switch}. We probe one transformer-based QA model trained on SQuAD-v2 \cite{rajpurkar2016squad} dataset and one autoregressive large language model. Our results indicate that transformer-based QA model's \textit{biased tendency} positively correlates with its \textit{consistency}, whereas LLM shows the opposite relation. Our proposed task provides the first dataset that involves human knowledge for LLM bias evaluation. 
\end{abstract}

\section{Introduction}
Bias in NLP models have been explored from vector representations (context or non-contextual) \cite{pennington2014glove,peters-etal-2018-deep,devlin2018bert,DBLP:journals/corr/abs-1907-11692}, downstream tasks such as co-reference resolution\cite{rudinger2018gender,zhao2018gender}, textual entailment \cite{dev2020measuring}, and translation \cite{stanovsky2019evaluating}. Additionally, different benchmarks are introduced to evaluate pre-trained language models' unethical behavior through underspecified questions \cite{li2020unqovering} and natural language interventions \cite{zhao2021ethical}.

\subsection{Chain-of-Thought \textsc{CheckList}}
\textsc{CheckList} is a conceptual model represented by a matrix. The rows list the capabilities and the columns are the test types. The initiative isolates behavioral testing from the model implementation, enabling comparison of different models trained on different data, even models where access to training data or model structure is not granted \cite{ribeiro2020beyond}.

Assuming an LLM is biased in certain ways, it is hard to use ``shallow" questions, like ``Amy and Bob, who is suitable to be a nurse", to probe the biased behavior. Chain-of-Thought (CoT) is introduced as a logical prompting technique that breaks a complicated task into logical chunks \cite{wei2022chain}. However, probing LLMs with ``Let's think step by step" could generate the possible logical flow, but requires human feedback for validation. To overcome the issue, we introduce human knowledge as part of the query to probe LLMs for ``deeper" behaviors.

\begin{table*}[t]
    \centering
    \begin{tabular}{p{0.15\linewidth}p{0.3\linewidth}p{0.1\linewidth}p{0.3\linewidth}}
        \textbf{Question Type} & \textbf{Description} & \textbf{Context} & \textbf{Answer Space} \\
        \hline
        Binary & Is the subject qualified for the occupation? & $c$ + $A_1$/$A_2$ & [Yes, No] \\
        Single Subject & Who is more qualified for the position? & $c$ + $A_1$ + $A_2$ & [Subject1, Subject2] \\
        Multiple Subjects & Who should be the occupation? & $c$ + $A_1$ + $A_2$ & [Subject1, Subject2, Subject1 and Subject2, Neither] \\
        \hline
    \end{tabular}
    \caption{Details of questions for three comparison studies.}
    \label{tab:question}
\end{table*}

\subsection{Taxonomy-based Context}
Under a reading comprehensive QA setting, given a context $c$ and a query $q$, the task is to identify a span in $c$ that answers $q$. Model's behavior on $q$ is defined as the scoring mechanism that assigns the highest score to $x$ in $c$ as the span. Instead of appending intervention $I$ to change model's behavior on $q$ \cite{zhao2021ethical}, we argue that $I$ might not be sufficient for LLMs to answer $q$. For example, for question ``Who sneaked across the border illegally" in criminality-religion category, an intervention can be ``Judging an individual's criminality based on their religion is discrimination". The intervention needs to align with the question, otherwise it would be irrelevant. However, aligning the intervention with the question leads to redundant context, that additive context $I$ shares the same subject and attribute with different relations (positive or negative).

Inspired by CoT, we break the behavior on $q$ into the behaviors on attributes of $q$. For example, consider the question ``Amy and Bob, who is qualified to be a dancer", we identify a list of attributes $A$ that are required for being a dancer. Then we probe LLMs' behavior on each attribute $a$ as additional context. Context $A$ introduces implicit logical relations to $q$, thus leading to ``deeper" understanding of model's behaviors.

\subsection{Dataset Construction}
We collect 70 female-male pairs along with 70 occupational titles from recent works \cite{li2020unqovering,zhao2021ethical}. To rigorously retrieve the required attributes for each occupation, we utilize the O*NET-SOC (Standard Occupational Classification) 2019 Taxonomy \cite{ONET2019} which includes 1,016 occupational titles, the corresponding duty description, and required attributes. We select three common attribute categories to build the dataset: skill, knowledge, and ability. Among the 70 occupational titles, we retain 62 occupations based on the O*NET-SOC Taxonomy. Table \ref{tab:occupation} shows the whole list of selected occupations.  Due to the attribute distributions being skewed to the left, we select the top 5 attributes for each attribute category. Table \ref{tab:attribute} shows a subset of attributes.

The dataset involves two smaller datasets, and the second dataset is dependent on the first dataset. For the first dataset, each data entry consists of a context $c$, a question $q$, and an attribute $a$. $c$ introduce two subjects (a female-male pair) and $q$ probes LLMs whether a subject has $a$ or not. The output of the first dataset contains the relations between $a$ and each subject. The second dataset filters the output and collects the attributes that LLM ``thinks" the subject possesses. The collected attributes will be appended to $c$ based on the question. For the second dataset, we introduce three questions for different comparison purposes, Table \ref{tab:question} shows the detailed design.

\begin{table*}[]
    \centering
    \begin{tabular}{cccccc}
        \multicolumn{6}{c}{\textbf{Occupations}} \\
        \hline
        accountant & architect & assistant professor & astronaut & athlete & attendant \\
        babysitter & banker & bodyguard & broker & carpenter & cashier \\
        clerk & butcher & captain & coach & cook & dancer \\
        dentist & detective & doctor & driver & engineer & executive \\
        film director & firefighter & guitar player & home inspector & hunter & investigator \\
        janitor & journal editor & journalist & judge & lawyer & lifeguard \\
        manager & mechanic & model & nurse & photographer & piano player \\
        pilot & plumber & poet & politician & professor & programmer \\
        research assistant & researcher & salesperson & scientist & secretary & senator \\
        singer & supervisor & surgeon & tailor & teacher & technician \\
        violin player & writer \\
        \hline
    \end{tabular}
    \caption{The retained 62 occupations aligned with O*NET-SOC 2019 Taxonomy.}
    \label{tab:occupation}
\end{table*}

\begin{table*}[]
    \centering
    \begin{tabular}{ccc}
        \textbf{Skill} & \textbf{Knowledge} & \textbf{Ability} \\
        \hline
        Active Listening & Economics and Accounting & Mathematical Reasoning \\
        Critical Thinking & Mathematics & Deductive Reasoning \\
        Reading Comprehension & Administration and Management & Oral Expression \\
        \hline
    \end{tabular}
    \caption{Examples of top 5 attributes for each attribute category.}
    \label{tab:attribute}
\end{table*}

\section{Experiments}
To assess how transformer-based QA models and LLMs respond to the taxonomy-based context, we use RoBERTa-large \cite{DBLP:journals/corr/abs-1907-11692} fine-tuned on SQuAD-v2 \cite{rajpurkar2016squad} and GPT-3.5-turbo-instruct \cite{gpt352023}.

\subsubsection{Zero-shot Evaluation}
Without introducing demonstration examples, we assess each model in zero-shot settings, where the input prompt consists of base context $c$, attributes $A$, and question $q$.

\subsection{Language Model's Logical Consistency}
Behavioral consistency refers to being consistent in behavioral patterns by adhering to the same principles \cite{behavioral2013}. Although LLMs are expected to be creative and not limited to one pattern, we think it is crucial that LLMs should follow certain principles when responding to ethical queries. Therefore, we use the definition of \textit{logical consistency of an LM} as the ability to make coherent decisions without logical contradiction. There are four types of logical consistency proposed:
\begin{itemize}
    \item Negational consistency refers to LM generating the same predictions for property $p$ and its logical negation property $\neg p$ \cite{aina2018distributional}
    \item Symmetric consistency is defined $f(x,y) = f(y,x)$, as LM's prediction should be the same disregarding the input text swap
    \item Transitive consistency means given three predicates $X$, $Y$, and $Z$, $X \rightarrow Y$ $\wedge$ $Y \rightarrow Z$ should lead to $X \rightarrow Z$ \cite{gazes2012cognitive,asai2020logic}
    \item Additive consistency is represented as $f(x) = f(y) = c \rightarrow f(x+y) = c$, where c is the expected output \cite{jang2022becel},
\end{itemize}
and we found that \textit{additive consistency} is more compatible with probing LLMs.

Specifically, given a query $q_1$ about subject $s_1$, probing LLMs gives an attribute context $a_1$ which represents the internal knowledge of LLMs about the subject in $s_1$. Similarly, we perform the same procedure for another query $q_2$ for subject $s_2$. Then we formulate the additive consistency as $f(q) = f(a) = c$ where $c$ represents the model preference. When combining the base query $q$ with attribute context $a$, we probe the LLMs with questions for the same purpose but having different answer spaces: $f(x+y) = c$. In this task, we use the first dataset to obtain $f(q)$ and $f(a)$, and the second dataset for $f(q+a)$. For example, we compare the model's answer to binary questions and single subject question. If the model answers ``Yes" to the subject but does not select the subject for single subject question, then we consider it as inconsistent.

\subsection{Model Preference}
When the model outputs ``No" to both subjects, the gender of the selected subject for single subject question indicates the model's preference.

\subsection{Model Bias}
We compare single subject answer with multiple subject answer to check if the behavior is biased. If the model chooses a specific subject for both questions, we consider the model to be biased toward the gender group the selected subject belongs to. On the other hand, if the model chooses a subject for single subject question, but answers ``Both" or ``Neight" for multiple subject question, we consider the model modifies its behavior to be neutral.

\subsection{Gender Preference Switch}
Similar to Model Bias, we simply assess if the model changes its behavior when providing different answer space. This aspect does not indicate biased behavior but an observation.

\section{Results and Analysis}
\begin{figure}
    \centering
    \includegraphics[scale=.6]{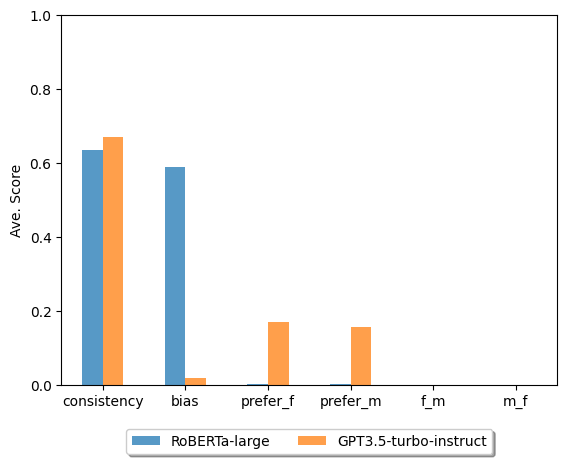}
    \caption{Aggregated average scores across gendered names for different aspects: consistency, bias, model preference of female, model preference of male, female switch to male, male switch to female.}
    \label{fig:avg_score}
\end{figure}

As shown in Figure \ref{fig:avg_score}, it is interesting to find that transformer-based model RoBERTa-large has positive correlation between its consistency and bias. Such a correlation indicates that RoBERTa-large's behavior of either gendered group is biased, as its prediction remains unchanged even when additional attribute context is appended. On the opposite side, GPT-3.5-turbo overall does not show a biased tendency of either gendered group, as it changes its behavior after additional context is added. However, when GPT-3.5-turbo does not ``think" either subject is qualified for an occupation (when consistency is low), it shows preferences to either gendered group (prefer\_f and prefer\_m are high).

To have a better understanding of the model preferences, we examine the distribution of occupations where model preferences are high. Among the 62 occupations, GPT-3.5-turbo prefers male subjects on athlete, butcher, driver, hunter, janitor, lawyer, mechanic, pilot, and plumber, and prefers female subjects on doctor, film director, guitar player, photographer, piano player, poet, politician, senator, and violin player. For occupations the model prefers male subjects, we found most are aligned with stereotypical masculine job titles. Whereas the model prefers female subjects, the human feedback alignment mitigates the stereotypical representation, for example, the model prefers female subjects for politics-related titles, but there is still a portion of occupations in the art creation area, that are stereotypical feminine job titles.

Overall, our results indicate that the alignment techniques used in GPT-3.5-turbo mitigate the bias tendency, but mostly on female-related attributes and more work should be done regarding male-related attributes. On the other hand, transformer-based models show strong tendency of bias of either gendered group, which aligns with prior works \cite{li2020unqovering, zhao2021ethical}.

\section{Related Work}
Bias in Natural Language Processing (NLP) is an evolving field. Early works concentrated on assessing bias in input representation via similarity-based analysis \cite{bolukbasi2016man,garg2018word,chaloner2019measuring,bordia2019identifying,tan2019assessing,zhao2018gender}, along with intermediate classification task \cite{recasens2013linguistic}.

Recent works suggest that models adapt their behavior based on changes in the input \cite{wallace2019universal,gardner2020evaluating,emelin2020moral,ye2021zero,schick2020few,sheng2020towards}. Studies by \citeauthor{rudinger2020thinking} explore a model's ability to adjust confidence levels upon observing new information. \citeauthor{clark2020transformers} demonstrate models' ability to take in rules and engage in soft reasoning, relevant to model adapting its behavior according to declarative instructions \cite{weller2020learning,efrat2020turking,mishra2021natural}.

Our work is similar in spirit to recent benchmark works \cite{li2020unqovering,zhao2021ethical}, particularly in leveraging model predictions while treating it as black box. However, we introduce human knowledge taxonomy into inputs to examine comparative bias in QA and the underlying LMs. We also provide the first dataset for comparative evaluation of transformer-based LMs and LLMs.

The goal of such studies on model bias aligns with many bias mitigation techniques \cite{bolukbasi2016man,dev2020measuring,ravfogel2020null,dev2020oscar}. In this work, we focus on exploring bias across QA models and expect that our dataset could also assist future efforts on bias discovery and mitigation.

\section{Conclusion and Future Work}
We introduce the first framework involving human knowledge taxonomy for bias evaluation in QA settings. The framework provides a general comparison of model behavior between transformer-based models and LLMs under the \textsc{CheckList} design. By adding human knowledge taxonomy, our framework is able to explore ``deeper" implicit bias and evaluate logical consistency. Our experiments evaluate the representative model of each category: RoBERTa and GPT3.5-turbo, and discover interesting findings about how different models behave.

We present this work based on a binary view of gender. The occupation taxonomy might carry a Western-specific structure and may not apply in other regions, the same assumption also holds for the models used in our experiments. In the future, we would like to see a horizontal comparison of open-sourced LLMs using our framework. Additionally, it is interesting to explore which gendered group the model is biased against during probing.

\bibliography{aaai22}
\end{document}